\documentclass[conference]{IEEEtran}

\usepackage{amsmath,amssymb}
\usepackage{graphicx}
\usepackage{booktabs}
\usepackage{array}
\usepackage{multirow}
\usepackage{url}
\usepackage[hidelinks]{hyperref}

\title{ROI-Driven Foveated Attention for Unified Egocentric Representations in Vision--Language--Action Systems}

\author{\IEEEauthorblockN{Xinhai Sun\IEEEauthorrefmark{1}\IEEEauthorrefmark{2}, Xiang Shi\IEEEauthorrefmark{1}, Menglin Zou\IEEEauthorrefmark{1}, Wenlong Huang\IEEEauthorrefmark{1}}
\IEEEauthorblockA{\IEEEauthorrefmark{1}Synthoid.ai, Shanghai, China \\
\IEEEauthorblockA{\IEEEauthorrefmark{2}Politecnico di Milano, Milan, Italy}}
Email: sun.xinhai@synthoid.ai, xinhai.sun@gsom.polimi.it}
\begin{document}
\maketitle

\begin{figure}[!t]
\centering
\setlength{\tabcolsep}{0pt}
\begin{tabular*}{\columnwidth}{@{\extracolsep{\fill}}ccc@{}}
\includegraphics[width=0.315\columnwidth]{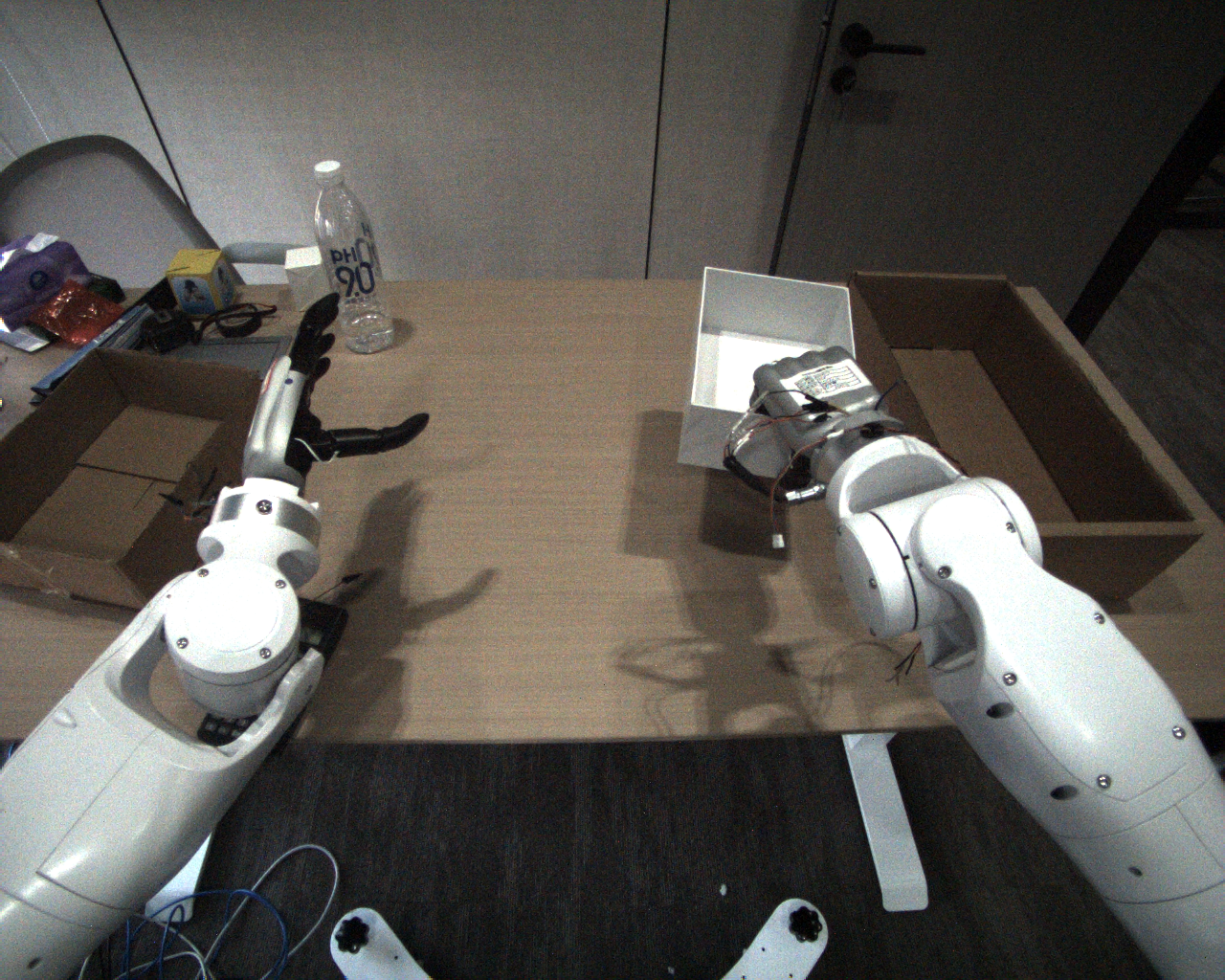} &
\includegraphics[width=0.315\columnwidth]{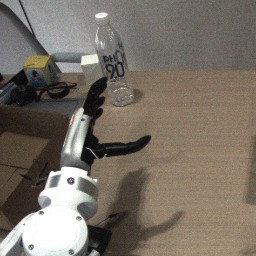} &
\includegraphics[width=0.315\columnwidth]{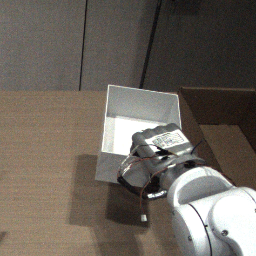} \\
\multicolumn{3}{c}{\parbox{0.96\columnwidth}{\centering\scriptsize Global-view observations and inward-offset hand-centric ROI crops during grasping, illustrating in-frame extraction under normal visibility.}} \\
\includegraphics[width=0.315\columnwidth]{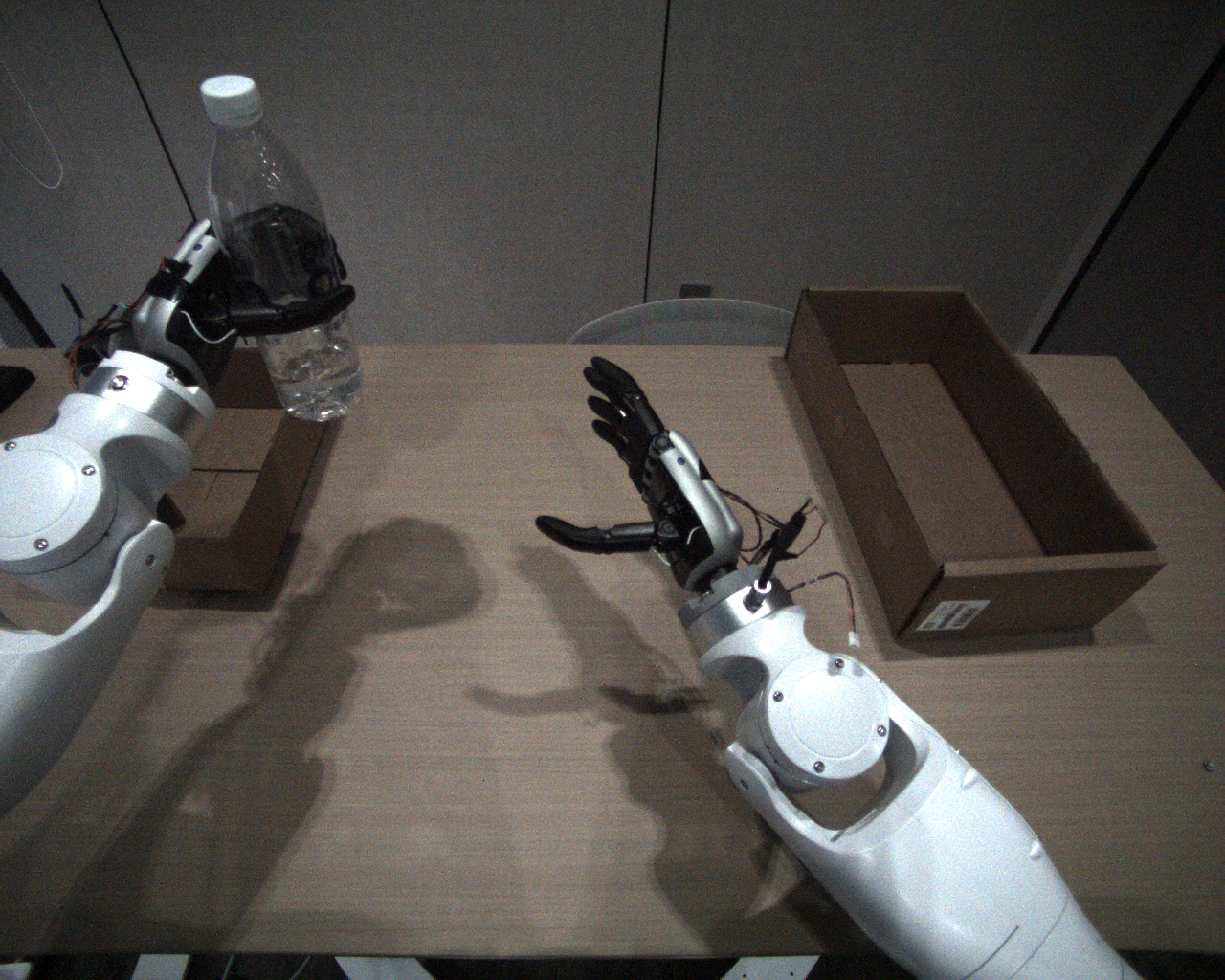} &
\includegraphics[width=0.315\columnwidth]{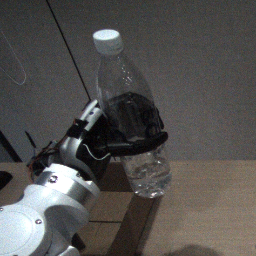} &
\includegraphics[width=0.315\columnwidth]{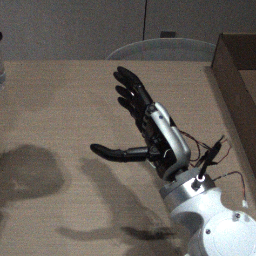} \\
\multicolumn{3}{c}{\parbox{0.96\columnwidth}{\centering\scriptsize Progressive hand--object interaction with ROI tracking the end-effector across motion, showing stable alignment and high-resolution capture of manipulation-relevant details.}} \\
\includegraphics[width=0.315\columnwidth]{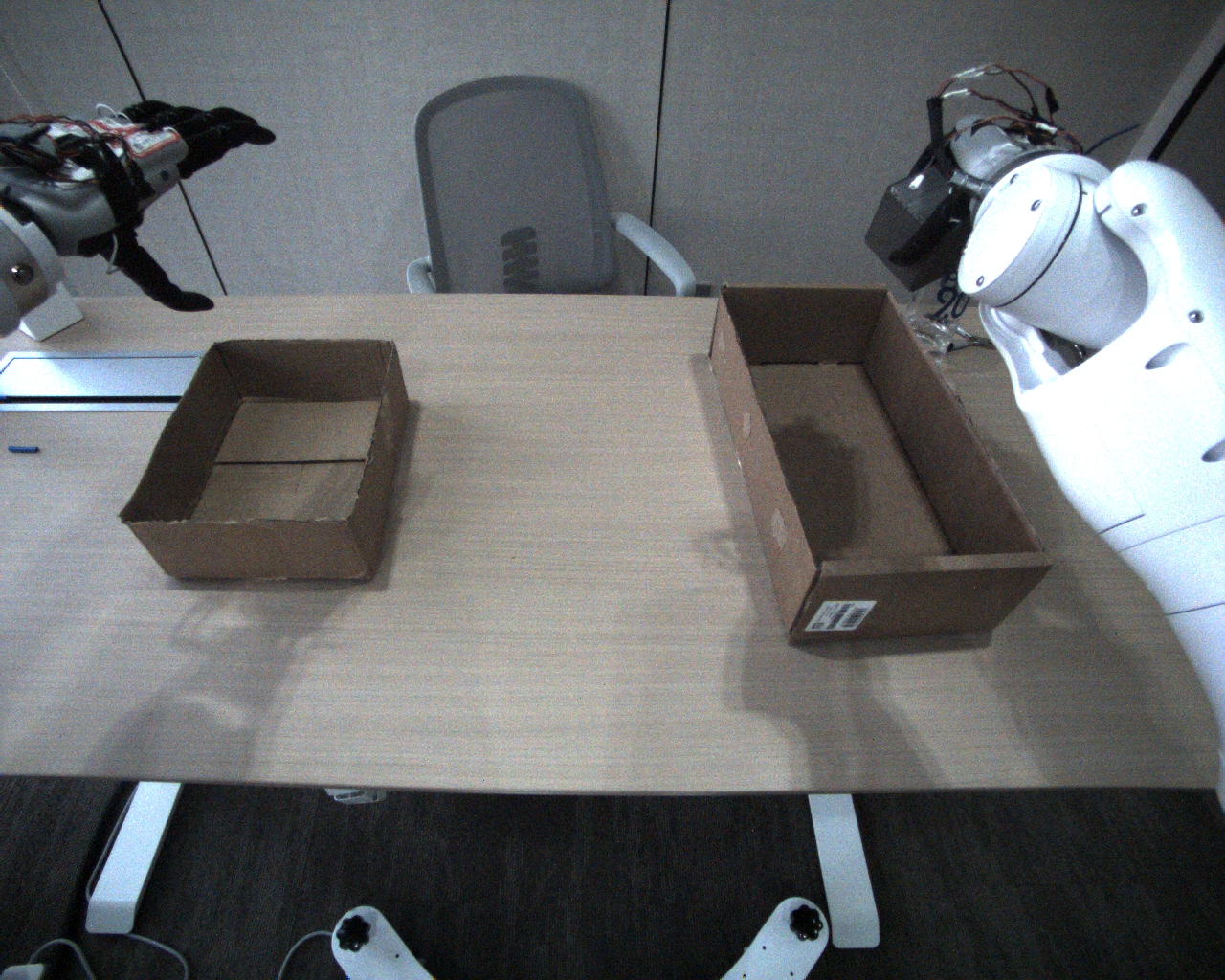} &
\includegraphics[width=0.315\columnwidth]{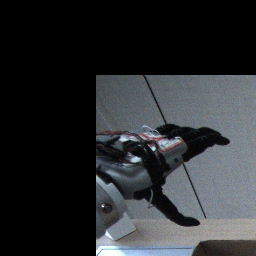} &
\includegraphics[width=0.315\columnwidth]{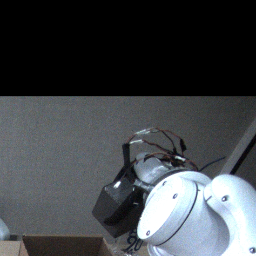} \\
\multicolumn{3}{c}{\parbox{0.96\columnwidth}{\centering\scriptsize Out-of-frame or partially visible end-effector cases where ROI crops extend beyond image boundaries and are completed by zero padding, ensuring consistent input size.}}
\end{tabular*}
\caption{ROI-based visual observations during manipulation.}
\label{fig:first_page_matrix}
\end{figure}

\begin{abstract}

The development of embodied AI systems is increasingly constrained by the availability and structure of physical interaction data. Despite recent advances in vision-language-action (VLA) models, current pipelines suffer from high data collection cost, limited cross-embodiment alignment, and poor transfer from internet-scale visual data to robot control. 

We propose a region-of-interest (ROI) driven engineering workflow that introduces an \emph{egocentric, geometry-grounded data representation}. By projecting end-effector poses via forward kinematics (FK) into a single external camera, we derive movement-aligned hand-centric ROIs without requiring wrist-mounted cameras or multi-view systems. Unlike directly downsampling the full frame, ROI is cropped from the original image before resizing, preserving high local information density for contact-critical regions while retaining global context.

We present a reproducible pipeline covering calibration, synchronization, ROI generation, deterministic boundary handling, and metadata governance. The resulting representation is embodiment-aligned and viewpoint-normalized, enabling data reuse across heterogeneous robots. We argue that egocentric ROI serves as a practical data abstraction for scalable collection and cross-embodiment learning, bridging internet-scale perception and robot-specific control.

\end{abstract}


\section{Introduction}
Large-scale embodied learning increasingly relies on cross-embodiment datasets to improve generalization and reuse demonstrations across platforms. This motivation is also reflected in our team's recent VLA system, SaiVLA-0 \cite{saivla0}. Recent large-scale efforts show that standardized data formats can support transfer across robot embodiments, but they also expose practical collection costs in real deployments \cite{du2023openx,brohan2022rt1}. In particular, DROID-style pipelines document substantial logistics for diverse real-world trajectories, including multi-view cameras, calibration assets, depth channels, and language annotations \cite{droid}.

At the same time, high-precision manipulation depends disproportionately on hand/tool-local visual evidence: contact transitions, subtle pose changes, and object-relative micro-motion \cite{levine2016end,kalashnikov2018qt}. A common response is to add wrist cameras (eye-in-hand) or more exocentric cameras. However, wrist cameras introduce recurring hand--eye calibration and maintenance burden, while multi-view setups amplify synchronization, bandwidth, and extrinsic calibration complexity. Hand--eye calibration itself is a classical AX=XB estimation problem, with a long line of methods (e.g., Tsai--Lenz and dual-quaternion formulations), underscoring its non-triviality in practice.

This paper proposes an engineering alternative: generate wrist/hand ROIs from a single external main camera via robot FK and calibrated projection, rather than capturing separate wrist-camera streams. In our VLA workflow, end-effector poses are projected into image coordinates to produce geometry-tied and movement-stabilized wrist crops with embodiment-aware center offsets; ROI inputs are always retained, and out-of-frame regions are deterministically zero-padded to a fixed resolution.

Our goal is not to propose a new model architecture or claim new benchmark results. Instead, we systematize data collection and training practices around FK-projected ROI so cross-robot teams can (i) collect consistent hand-centric supervision, (ii) reduce sensor and calibration burden, and (iii) standardize derived artifacts for transfer and reproducibility. This framing targets information-technology conference priorities: workflow design, data governance, and operational cost.

\section{Contributions:}
\begin{itemize}
    \item \textbf{A deterministic FK-to-ROI pipeline for egocentric perception alignment.} We introduce a geometry-driven mapping from robot FK to image-space ROI with explicit coordinate conventions and embodiment-aware inward center offsets. Unlike heuristic or learned attention selectors, the mapping is deterministic and reproducible across time, enabling stable manipulation-centric observation extraction. We further include fixed-size ROI construction with boundary-aware zero padding and confidence logging, turning ROI from a model-specific feature into a system-level abstraction.

    \item \textbf{A metadata and governance schema for reproducible ROI-derived artifacts.} We define ROI-related artifacts as first-class data products, including calibration parameters, projection metadata, temporal alignment, and versioned dependencies. This supports deterministic regeneration, validation, and sharing of ROI-augmented datasets, extending feature-caching workflows into a governance-aware data pipeline. From an information-systems perspective, the schema improves reproducibility, auditability, and cross-team interoperability under heterogeneous sensor setups.

    \item \textbf{A retrofit-oriented evaluation protocol without native ROI support.} We provide an evaluation protocol that reconstructs ROI views offline from logged robot states and calibration metadata, allowing controlled comparisons between baseline and ROI-augmented pipelines. This lowers adoption cost by avoiding new sensor instrumentation and supports systematic analysis of training efficiency, attention allocation, and control robustness. More broadly, it enables incremental upgrades of legacy datasets into ROI-aware representations.
\end{itemize}

\section{Related Work}
\subsection{Cross-Embodiment Manipulation Data}
Open X-Embodiment and DROID provide large-scale manipulation datasets, while RT-X is a model family trained on heterogeneous robot data from Open X-Embodiment dataset and additional curated datasets \cite{du2023openx,brohan2022rt1,droid}. Together, these efforts show that scale and standardization can improve transfer across robot embodiments, while also exposing the engineering cost of broad real-world coverage, calibration quality control, and multi-view data management.

\subsection{Hand-Centric and Egocentric Representations}
Egocentric corpora such as Ego4D and EPIC-KITCHENS show that hand--object interactions carry dense task-relevant information \cite{ego4d,epic}. In robotics, manipulation systems such as CLIPort, Deep Visual Foresight, and Transporter Networks similarly emphasize action-relevant local structure \cite{shridhar2023cliport,finn2017deep,zeng2020transporter}. This motivates pipelines to preserve hand-centric evidence while controlling capture complexity.

\subsection{Foveation and Attention}
Foveated perception and recurrent attention suggest that high-resolution processing of task-relevant regions can improve efficiency and stability \cite{mnih2014,patney2016foveated}. Recent VLA and multimodal robotics models (e.g., RT-1, RT-2, and PaLM-E) further motivate structured visual emphasis mechanisms for robust control grounding \cite{driess2023palm,posa_vla}.

\subsection{Calibration Foundations}
Classical hand--eye calibration methods, including Tsai--Lenz and dual-quaternion formulations, formalize camera--robot frame estimation and reveal practical sensitivity to setup quality \cite{tsai_lenz,daniilidis}.

\section{Method: FK-Projected ROI Generation}
\subsection{Assumptions and Coordinate Frames}
We assume: (1) joint encoder logs and a kinematic model, (2) at least one fixed external RGB camera, and (3) synchronized timestamps. We define robot base frame $\{B\}$, end-effector frame $\{E\}$, camera frame $\{C\}$, and image-plane pixel coordinates.

\subsection{FK-to-ROI Pipeline}
For each time step $t$:
\begin{enumerate}
    \item Compute end-effector pose from joint state:
    \begin{equation}
    {}^{B}\mathbf{T}_{E}(t)=\mathrm{FK}(\mathbf{q}_t)\in SE(3).
    \end{equation}
    where $\mathbf{q}_t$ denotes the robot joint configuration (joint-angle vector) at time step $t$.
    \item Transform to camera frame using calibrated eye-to-hand extrinsics:
    \begin{equation}
    \begin{bmatrix} {}^{C}\mathbf{p}_{E}(t) \\ 1 \end{bmatrix} =
    {}^{C}\mathbf{T}_{B}
    \begin{bmatrix} {}^{B}\mathbf{p}_{E}(t) \\ 1 \end{bmatrix}.
    \label{eq:extrinsic}
    \end{equation}
    \item Project to image pixels via the pinhole model:
    \begin{equation}
    \tilde{\mathbf{p}}(t)=\mathbf{K}\,[\mathbf{I}_{3\times3}\ \mathbf{0}]
    \begin{bmatrix} {}^{C}\mathbf{p}_{E}(t) \\ 1 \end{bmatrix},
    \quad
    (u,v)=\left(\frac{\tilde{p}_x}{\tilde{p}_z},\frac{\tilde{p}_y}{\tilde{p}_z}\right).
    \label{eq:projection}
    \end{equation}
    \item Apply an embodiment-aware center offset before cropping:
    \begin{equation}
    \begin{bmatrix} u_c \\ v_c \end{bmatrix}
    =
    \begin{bmatrix} u \\ v \end{bmatrix}
    + \beta\,\mathbf{d}_{\mathrm{in}}(t),
    \label{eq:offset_center}
    \end{equation}
    where $\mathbf{d}_{\mathrm{in}}(t)$ is the image-plane unit direction pointing from wrist center toward the manipulator inner side (typically fingertip/contact side), and $\beta$ is an embodiment-specific offset magnitude in pixels.
    \item Crop ROI centered at $(u_c,v_c)$ using fixed or depth-adaptive scaling, and apply zero padding when the crop extends beyond image boundaries.
    \item Compute ROI confidence as metadata for monitoring and analysis; all frames are retained for training and inference without confidence-based ROI replacement.
\end{enumerate}

To ensure reproducible depth-adaptive cropping, we define a reference ROI scale as:
\begin{equation}
\ell(t)=\mathrm{clip}\!\left(\alpha\,\frac{f_x\,r}{Z_c(t)},\ \ell_{\min},\ \ell_{\max}\right),
\end{equation}
where $f_x$ is the focal length in pixels, $r$ is a nominal physical ROI radius, $Z_c(t)$ is end-effector depth in the camera frame, and $(\ell_{\min},\ell_{\max})$ enforce practical crop bounds.

In practice, ROI is often cropped as an adaptive rectangular window rather than a strictly square patch, so the crop can better match gripper morphology and tool geometry. The horizontal and vertical crop bounds are adjusted using embodiment-specific geometry priors while remaining centered on the offset target $(u_c,v_c)$. This offset avoids excessive forearm/wrist background when the FK end-effector origin is not colocated with the visually salient contact region. Each cropped ROI is then resized to a fixed input resolution of $256\times256$ using interpolation (e.g., area-based resampling such as \texttt{INTER\_AREA}), ensuring shape-consistent model input while preserving fine manipulation cues.

For bimanual systems, the pipeline is applied independently to each arm.

\subsection{Why This Lowers Operational Burden}
Compared with wrist cameras, this workflow removes camera mounting/wiring on the end-effector and avoids recurring eye-in-hand recalibration loops. Compared with multi-view capture, it preserves a hand-centric channel while limiting sensor count and synchronization overhead.

\subsection{Why ROI Reduces Sensor and Calibration Burden}
Eye-in-hand setups require estimating and repeatedly maintaining the camera-to-end-effector transform $H^{EE}_{CAM}$ after hardware or tool changes. Our ROI workflow uses a fixed external camera and FK-projected wrist crops, reducing recurring overhead:
\begin{itemize}
    \item no wrist-camera mounting/wiring and reduced mechanical maintenance,
    \item no per-tool recalibration loop and no extra wrist-stream synchronization path.
\end{itemize}

ROI also reduces dependence on dense multi-view rigs. While multi-view capture improves observability, it increases camera count, calibration workload, and synchronization complexity. By projecting FK to the primary image and cropping a hand-centric window, one camera can provide both global context and a derived local channel.

\textbf{Practical clarification.} ROI projection still requires calibration: camera intrinsics and camera-to-base extrinsics are necessary. This is an eye-to-hand step, but it is typically more stable operationally than repeatedly maintaining eye-in-hand transforms. Because even eye-to-hand extrinsics may drift across sessions, robust logging, versioning, and post-hoc validation remain required.

\section{Engineering Workflow for Camera-Ready Deployment}
We treat ROI as a deterministic \emph{derived artifact} generated from canonical logs. This shifts complexity from hardware to versioned software transformations.

\subsection{Pipeline Stages}
\begin{enumerate}
 \item \textbf{Setup}: Define a unified frame convention (robot base, end-effector, and camera frames), and register camera intrinsics and camera-to-base extrinsics with explicit versioning. Establish a time-synchronization policy (e.g., hardware trigger or software timestamp alignment) to ensure consistent pairing between image frames and robot states.

    \item \textbf{Collection}: Record synchronized RGB streams and robot-state trajectories (joint positions, end-effector poses) at matched frequencies. Log raw timestamps, sensor identifiers, and buffering delays to enable post-hoc temporal alignment and auditability.

    \item \textbf{Offline generation}: Compute forward kinematics (FK) to obtain end-effector poses in the base frame, project them into image coordinates using calibrated intrinsics/extrinsics, and generate ROI crops via deterministic crop-and-resize operations. Augment each ROI with confidence scores derived from projection validity, visibility, and boundary conditions.

    \item \textbf{Validation}: Perform reprojection consistency checks (FK $\rightarrow$ image $\rightarrow$ ROI), compute coverage and in-frame ratios, and screen for temporal drift or discontinuities. Aggregate statistics such as projection success rate and ROI stability to identify calibration or synchronization issues.

    \item \textbf{Packaging}: Organize outputs into structured manifests with explicit schema definitions, including metadata fields, calibration versions, and ROI descriptors. Attach checksums and version tags for reproducibility, and enforce access control policies for dataset sharing and governance.

    \item \textbf{Training/transfer}: Integrate ROI streams with global-view inputs via fusion mechanisms (e.g., concatenation or cross-attention), and enable per-robot regeneration of ROI data using the same FK-to-projection pipeline. This supports cross-embodiment transfer without requiring re-collection of raw sensory data.
\end{enumerate}

\subsection{Recommended Metadata Fields}
For each frame, we recommend a compact metadata record that acts as a data contract for ROI reconstruction, validation, and governance. At minimum, it should include four field groups: temporal/identity fields (timestamp, camera ID, robot ID, arm ID) for traceability; geometric descriptors (projected center, inward-offset center, crop size, ROI confidence, in-frame ratio, padding ratio) for spatial definition and reliability; kinematic/projection context (end-effector pose in base frame and projected point in camera frame) for geometry-grounded reconstruction; and provenance/version fields (intrinsics version, extrinsics version, ROI-generator version, crop-offset policy version, padding mode, quality flags) for lineage.

These fields provide three system-level functions. First, they ensure \emph{deterministic regeneration}: with frame geometry and versioned calibration, ROI streams can be reproduced across runs and sites. Second, they support \emph{fine-grained failure diagnosis}: confidence, padding statistics, and quality flags reveal causes such as occlusion, temporal misalignment, calibration drift, or kinematic inconsistency. Third, they improve \emph{governance and interoperability}: explicit identifiers and version metadata make ROI outputs auditable, shareable, and easier to integrate across platforms.

\subsection{Quality Gates}
To ensure reliability and consistency, we define quality gates as validation checkpoints for both data generation and consumption. Time-alignment residuals quantify synchronization error between image and robot-state streams. Projection validity ratio measures the share of frames with valid in-frame FK projections. ROI coverage checks whether the crop captures the end-effector interaction region without truncation or drift. Temporal stability (jitter) tracks frame-to-frame variation in ROI position and scale. Padding ratio and in-frame ratio quantify partial observability and out-of-frame extent, serving as practical diagnostics for occlusion, calibration degradation, or domain shift.

Beyond validation, these gates act as continuous system-health signals. In large-scale or cross-site deployments, they offer early warnings of pipeline degradation and enable automated alerting, filtering, or recalibration.

\section{Training and Annotation Practices}
\subsection{ROI as Foveal Supervision}
ROI acts as a high-acuity local channel for contact-centric cues, while the full frame preserves scene-level context.

A practical advantage is information-density preservation. Let a raw frame be downsampled from $W_0\times H_0$ to $W_g\times H_g$ for global input, with scale factor $s=W_g/W_0\approx H_g/H_0$. A local region of width $w_{roi}$ in the raw frame is represented by only $s\,w_{roi}$ pixels in the downsampled global image. If we instead crop this region first and then resize it to $W_r$, the effective local detail scale is
\begin{equation}
\rho = \frac{W_r}{s\,w_{roi}}.
\label{eq:density_gain}
\end{equation}
For a typical setting ($1280\rightarrow256$, so $s=0.2$) and $w_{roi}\approx256$, the same hand region occupies about $51$ pixels in the global branch but $256$ pixels in the ROI branch, i.e., roughly $5\times$ higher local sampling density.

This aligns with the foveation principle: only a small visual region carries high-acuity detail, while peripheral context supports global scene understanding. In practice, fixed-size ROI-main fusion with deterministic padding provides a robust default for real deployment.

\subsection{ROI-Induced Attention Forcing}
Although ROI regions are cropped from the same source image, we treat them as independent high-resolution inputs to explicitly bias the attention distribution of the vision transformer (ViT). Concretely, each observation is constructed as
\begin{equation}
\mathcal{I}=\{I_{\text{global}}, I_{\text{roi},1}, I_{\text{roi},2}\},
\quad I_* \in \mathbb{R}^{256\times256},
\end{equation}
where $I_{\text{global}}$ is the full-frame observation and $I_{\text{roi},i}$ are inward-offset, end-effector-anchored crops derived via FK projection.

Each image is independently patchified into tokens:
\begin{equation}
T_* = \text{PatchEmbed}(I_*) \in \mathbb{R}^{N\times d},
\end{equation}
and the full token sequence is formed by concatenation:
\begin{equation}
T = [T_{\text{global}};T_{\text{roi},1};T_{\text{roi},2}] \in \mathbb{R}^{(3N)\times d}.
\end{equation}

\subsection{Implicit Attention Reweighting}
In standard ViT self-attention,
\begin{equation}
\text{Attn}(Q,K,V)=\text{softmax}\!\left(\frac{QK^{\top}}{\sqrt{d}}\right)V,
\end{equation}
the attention allocation across spatial regions is determined by token similarity. By duplicating and isolating ROI regions as separate token groups, we introduce a structural bias:
\begin{itemize}
    \item ROI pixels are represented multiple times (in global and ROI views),
    \item ROI tokens occupy a larger fraction of the token space,
    \item ROI tokens have higher spatial resolution per semantic unit.
\end{itemize}

Let $T_{\text{roi}}\subset T$ denote ROI tokens. Their effective contribution to attention can be approximated as
\begin{equation}
\mathbb{E}[\text{Attn weight on }T_{\text{roi}}]
\propto
\frac{|T_{\text{roi}}|}{|T|}+\Delta_{\text{dup}},
\end{equation}
where $\Delta_{\text{dup}}$ captures redundancy-induced reinforcement from duplicated visual content.

\subsection{Effect as Structural Attention Forcing}
This design induces implicit attention forcing without modifying the transformer architecture:
\begin{itemize}
    \item \textbf{Token-level amplification}: ROI regions are over-represented in the token sequence, increasing their probability mass in attention computation.
    \item \textbf{Multi-scale consistency}: the same physical region appears in both global and zoomed-in views, encouraging cross-token alignment and reinforcing feature learning.
    \item \textbf{Inductive bias toward interaction regions}: because ROIs are end-effector-anchored with inward offsets, the model repeatedly sees manipulation-relevant details and improves fine-grained control sensitivity.
\end{itemize}

\subsection{Implications for VLA Systems}
From a VLA perspective, the input is not purely visual, but a multimodal tuple consisting of global view, ROI views, language instruction, and robot proprioceptive state:
\begin{equation}
x = \left(I^{global}, I^{roi_1}, I^{roi_2}, w, s \right)
\end{equation}
where $w$ denotes the language instruction and $s$ denotes the robot state (e.g., joint positions, gripper states, or end-effector states).

The visual stream is encoded as:
\begin{equation}
T^{vis} = [T^{global}; T^{roi_1}; T^{roi_2}]
\end{equation}
the language input is mapped to text tokens
\begin{equation}
T^{txt} = \text{TextEmbed}(w),
\end{equation}
and the robot state is projected into the same latent space as
\begin{equation}
T^{state} = \phi(s),
\end{equation}
where $\phi(\cdot)$ is a learnable state projection.

The final multimodal token sequence is then constructed as
\begin{equation}
T^{all} = [T^{vis}; T^{txt}; T^{state}],
\end{equation}
which is processed by the VLA backbone:
\begin{equation}
H = \text{Transformer}(T^{all}).
\end{equation}
where $H$ denotes the multimodal hidden representations output by the backbone.

Under this formulation, ROI does not replace the global image, language, or proprioceptive input; instead, it reshapes the visual token distribution inside the full multimodal context. The resulting effect is that hand-centric interaction regions receive stronger representation capacity while still remaining grounded in task instruction and robot embodiment.

This mechanism removes the need for explicit attention supervision, auxiliary losses, or architectural modifications, while still encouraging the model to allocate more representational emphasis to egocentric interaction regions.

\subsection{Architecture-Agnostic Training Practices}
Without introducing new model heads, attention to ROI can be strengthened through implementation-level training design:
\begin{itemize}
    \item \textbf{ROI/main-view dropout scheduling:} randomly drop or degrade the main view in a fraction of batches so the policy must rely on ROI for fine manipulation.
    \item \textbf{ROI-weighted learning windows:} upweight samples with high ROI confidence or likely contact phases (e.g., low end-effector velocity) to emphasize micro-motion modeling.
    \item \textbf{Pose-conditioned metadata supervision:} store projected center, crop scale, ROI confidence, and depth proxy for filtering, debugging, and optional weak supervision.
\end{itemize}

\subsection{Annotation Efficiency}
Because ROI centers are generated geometrically, manual hand-box labeling can be reduced and annotation consistency can improve across robots.

Kinematics-derived labels also improve temporal consistency: ROI follows end-effector motion deterministically, reducing discontinuities from detector drift and manual annotation variance.

\subsection{Cross-Robot Transfer Workflow}
For cross-embodiment deployment, we recommend a canonical ROI format and strict version control:
\begin{itemize}
    \item fixed crop size and normalization (e.g., $256\times256$),
    \item consistent naming/schema (arm id, projection center, confidence),
    \item deterministic version tuple: kinematics version, calibration version, and crop-policy version.
\end{itemize}
This is important because ROI is not a generic image transform; it is geometry-conditioned and therefore sensitive to FK and calibration integrity.

\subsection{Confidence-Aware Safety}
Low-confidence ROI states are not discarded; instead, the pipeline preserves fixed-size ROI inputs with explicit zero padding and logs confidence/padding metadata for monitoring and safety analysis. This preserves training--inference consistency while allowing the policy to rely on multimodal cues, including proprioceptive state, under partial observability.

\section{Evaluation Protocol Without Native ROI Streams}
This paper provides protocols rather than new experimental claims.

\subsection{Engineering Metrics}
Measure onboarding time, sensor count/bandwidth, calibration step count and failure rate, storage overhead, and reproducibility consistency.

\subsection{Geometric ROI Quality Metrics}
Measure ROI coverage, reprojection residual proxy, temporal jitter, padding ratio, in-frame ratio, and calibration drift indicators.

\subsection{Learning Proxy Metrics}
Measure data efficiency under fixed budgets, precision-control surrogate tasks, and action-stability indicators (e.g., jitter, jerk, action flips).

\subsection{Practical Evaluation Routes}
\textbf{Route A (Dataset Retrofit):} For datasets with calibration and robot states, generate ROI offline and run controlled training protocols. To reflect real VLA deployment across embodiments, include split settings that vary manipulator morphology, camera baseline, gripper/tool geometry, and control frequency. In addition to global-only vs. global+ROI comparisons, report robustness under action-space remapping (e.g., joint-space to Cartesian interfaces), latency injection, and partial calibration drift, which are common failure sources when transferring policies from one robot to another.

\textbf{Route B (Simulation/Replay):} Use simulation or deterministic replay to validate projection and boundary-padding logic before hardware trials. This route should explicitly stress cross-embodiment deployment gaps: kinematic mismatch, observation delay, asynchronous sensing, and contact-model differences. Recommended outputs are degradation curves and failure taxonomies that map directly to deployment decisions (recalibration, policy retuning, or conservative gating thresholds).

A concise comparison of practical trade-offs across hand-centric collection strategies is summarized in Table~\ref{tab:collection_comparison}.

\begin{table*}[t]
\caption{Comparison of Hand-Centric Data Collection Strategies}
\label{tab:collection_comparison}
\centering
\begin{tabular}{p{2.9cm}p{1.6cm}p{2.9cm}p{2.1cm}p{1.6cm}p{4.2cm}}
\toprule
\textbf{Strategy} & \textbf{Cameras} & \textbf{Calibration Burden} & \textbf{Sync Cost} & \textbf{Hand-Centric Signal} & \textbf{Notes} \\
\midrule
Single external camera only & 1 fixed & Low--Medium (eye-to-hand) & Low & Low--Medium & Simplest setup; weak local contact detail. \\
Wrist camera (eye-in-hand) & 1+ wrist & High & Medium & High & Strong local view; recurring hand--eye maintenance required. \\
Multi-view exocentric setup & 2--4+ fixed & High & High & Medium--High & Good observability; expensive calibration/synchronization. \\
Single camera + FK-projected ROI (ours) & 1 fixed & Medium (one-time eye-to-hand + validation) & Medium & High & Derived hand-centric stream without extra sensors; requires robust QC. \\
\bottomrule
\end{tabular}
\end{table*}

\section{Limitations and Governance}
\subsection{Limitations}
\textbf{Calibration and cross-embodiment geometry drift.} ROI quality depends on FK correctness, camera intrinsics/extrinsics, and time synchronization, and can further degrade when end-effectors or dexterous hands change across platforms. These changes alter wrist-to-tool geometry and may require different ROI center/scale policies, making pure FK-centered crops less consistently aligned and increasing representation drift risk. Therefore, post-hoc calibration checks, reprojection quality metrics, and distribution monitoring of ROI statistics (center, scale, in-frame ratio, padding ratio) are necessary safeguards.

\textbf{Occlusion and self-occlusion.} Hand-centric regions are frequently occluded by the manipulator, gripper, or manipulated objects. These time-varying occlusions can reduce ROI reliability exactly when contact dynamics become important. A practical mitigation is to keep ROI and global view jointly available while preserving deterministic padded ROI inputs and multimodal conditioning.

\textbf{Context loss from over-foveation.} Overweighting ROI may suppress global scene cues required for search, long-horizon planning, and multi-object reasoning. For this reason, ROI should be a complementary channel rather than a replacement of the main view.

\textbf{Ongoing ROI experimentation.} ROI-focused experiments are still in progress, including controlled ablations on crop policy, offset strategy, and cross-embodiment transfer. Therefore, this paper emphasizes a reproducible engineering workflow and governance schema rather than claiming final quantitative superiority.

\subsection{Governance Recommendations}
\textbf{ROI formation under closed-loop teleoperation.} Teleoperation logs should be modeled as closed-loop trajectories in which observation, action, and delay jointly shape ROI. Beyond frame-level geometry, ROI generation should be conditioned on control-state variables, including end-to-end latency $L_t$, effective control frequency $f_t$, and operator-to-robot mapping residual $r_t=\|g(u_t)-x_t^{cmd}\|$. In practice, we recommend attaching $(L_t,f_t,r_t)$ to each ROI record together with projection validity and padding ratio. This makes it possible to separate manipulation-relevant ROI from ROI patterns caused by delay compensation, oscillatory correction, or interface mismatch.

\textbf{Lifecycle traceability of ROI transformations.} Consistent with the paper's deterministic FK-to-ROI pipeline, governance should treat ROI as a versioned derived artifact rather than a transient image crop. Each stage---FK projection (Eqs.~(\ref{eq:extrinsic})--(\ref{eq:projection})), inward-offset center policy (Eq.~(\ref{eq:offset_center})), crop/resize, and padding---should be reproducible from metadata. We recommend a minimal lineage tuple: \{intrinsics\_ver, extrinsics\_ver, fk\_ver, roi\_generator\_ver, crop\_offset\_ver, resize\_policy, padding\_mode\}. This directly supports the conclusion goal of deterministic regeneration and auditable deployment.

\textbf{Cross-embodiment consistency of ROI representations.} ROI portability requires explicit cross-embodiment remapping, rather than direct reuse of pixel-space crops. The schema should minimally encode coordinate frames, camera-to-base transform, end-effector/tool geometry, control modality, and teleoperation mappings when applicable. Compatibility should be verified by tracking shifts in projected center, offset-center, crop scale, in-frame ratio, and padding ratio before/after remapping.

\textbf{ROI stability beyond human-in-the-loop control.} ROI trajectories stabilized by continuous teleoperation feedback may not remain stable in autonomous rollout. We therefore recommend feedback-reduction evaluations (delayed, intermittent, or no-feedback replay) with the same quality gates, while tracking ROI jitter, recovery latency, and failure modes. Segment-level labels separating nominal progression from intervention-driven correction further enable targeted stress tests for reliable autonomous deployment.

\section{Conclusion}

We presented a camera-ready engineering workflow for generating hand-centric supervision from a single external camera via FK-projected ROI derivation. By replacing multi-view or wrist-mounted sensing with a deterministic, kinematics-driven projection, the approach reduces system complexity while preserving manipulation-critical visual cues, shifting the focus from sensor design to structured data generation.

We formalized ROI as a reproducible derived artifact with explicit coordinate conventions, inward-offset crop policy, deterministic padding, and governance-aware metadata, enabling consistent regeneration, validation, and scalable data pipelines across heterogeneous robotic setups.

Furthermore, we argued that ROI induces an \emph{egocentric data paradigm}: a geometry-grounded representation anchored to the end-effector, producing embodiment-aligned, viewpoint-normalized, and interaction-centric observations. This facilitates representation unification across diverse data sources, including robot demonstrations, simulation, and human video, providing a practical pathway toward integrating internet-scale perception with robot control.

Although dedicated empirical validation remains future work, the proposed abstraction offers a principled and scalable direction for cross-embodiment and transfer-oriented VLA systems.

\end{document}